\pdfoutput=1

\documentclass[11pt]{article}

\usepackage[preprint]{acl}

\usepackage{times}
\usepackage{latexsym}
\usepackage{graphicx}  
\usepackage{float}    
\usepackage{hyperref} 
\usepackage{booktabs}
\usepackage[table]{xcolor}
\usepackage{colortbl} 
\usepackage{subcaption}
\usepackage{geometry}
\usepackage{graphicx}
\usepackage{pifont} 
\usepackage{pifont} 
\usepackage{booktabs}
\usepackage{geometry}
 \usepackage{multirow}

\usepackage[T1]{fontenc}

\usepackage[utf8]{inputenc}

\usepackage{microtype}

\usepackage{inconsolata}

\usepackage{graphicx}

\usepackage{amsmath,amsfonts,bm}

\def\eqref#1{equation~\ref{#1}}

\def\1{\bm{1}}

\DeclareMathAlphabet{\mathsfit}{\encodingdefault}{\sfdefault}{m}{sl}
\SetMathAlphabet{\mathsfit}{bold}{\encodingdefault}{\sfdefault}{bx}{n}

\usepackage{amsmath,amsfonts,bm}
\usepackage{booktabs} 
\usepackage{url}
\usepackage{graphicx}
\usepackage{array}
\usepackage{color}
\usepackage{makecell}
\usepackage{multirow}
\usepackage{xspace}
\usepackage{colortbl}
\usepackage{subcaption}
\usepackage{algpseudocode}
\usepackage[noend,ruled]{algorithm2e}
\usepackage{setspace}
\usepackage{varwidth}
\usepackage{enumitem}
\usepackage{mdframed}
\usepackage{centernot}
\usepackage{arydshln}

\usepackage{soul}
\usepackage{pifont}
\usepackage{graphics}

\usepackage{wrapfig}

\usepackage{cuted,tcolorbox,lipsum}

\newcommand{\cmark}{\textcolor{green!70!black}{\ding{51}}} 
\newcommand{\xmark}{\textcolor{red!80!black}{\ding{55}}} 

\newcommand{\modelname}{\textsc{SACL}\xspace}

\newcommand{\circled}[1]{\textcircled{\raisebox{-0.9pt}{#1}}}

\newcommand{\start}[1]{\vspace{.3mm}\noindent{{\bf #1}.}}

\definecolor{amber}{rgb}{1.0, 0.75, 0.0}
\definecolor{applegreen}{rgb}{0.55, 0.71, 0.0}
\definecolor{treegreen}{rgb}{0.13, 0.54, 0.23}
\definecolor{LightCyan}{rgb}{0.88,1,1}
\definecolor{LightBlue}{RGB}{171, 227, 235}
\newcommand{\bad}{\cellcolor{red!8}}
\newcommand{\worse}{\cellcolor{red!16}}
\newcommand{\worst}{\cellcolor{red!24}}
\newcommand{\good}{\cellcolor{applegreen!10}}
\newcommand{\better}{\cellcolor{applegreen!20}}

\definecolor{green2}{HTML}{BFD8B6}
\definecolor{green3}{HTML}{E7F0E5}
\definecolor{greenarrow}{HTML}{1DB100}
\definecolor{red3}{HTML}{C82506}

\definecolor{gred}{RGB}{255,102,102}
\definecolor{gblue}{RGB}{51,102,255}
\definecolor{gyellow}{RGB}{244,180,0}
\definecolor{ggreen}{RGB}{15,157,88}
\definecolor{ggrey}{RGB}{115,115,115}
\definecolor{na}{gray}{0.9}

\definecolor{textRed}{RGB}{157,0,23}
\definecolor{textYellow}{RGB}{166,119,54}
\definecolor{textGreen}{RGB}{58,110,38}
\definecolor{textBlue}{RGB}{39,71,156}

\definecolor{LightYellow}{RGB}{255,250,208}
\definecolor{LightGreen}{RGB}{194,255,192}
\definecolor{LightBlue}{RGB}{187,236,251}
\definecolor{LightPurple}{RGB}{224,223,255}
\definecolor{LightGrey}{RGB}{225,225,225}

\definecolor{Grey}{RGB}{150,150,150}

\definecolor{OrangeRed}{rgb}{1.0, 0.27, 0.0}
\definecolor{midnightgreen}{rgb}{0.0, 0.29, 0.33}
\definecolor{darkgreen}{rgb}{0.0, 0.42, 0.24}

\definecolor{diagramRed}{RGB}{246,193,193}
\definecolor{diagramPurple}{RGB}{224,224,253}
\definecolor{diagramOrange}{RGB}{244,222,176}

\title{\modelname: Understanding and Combating Textual Bias in Code Retrieval \\
with Semantic-Augmented Reranking and Localization
}

\author{Dhruv Gupta \\
  Carnegie Mellon University \\
  \texttt{dhruvgu2@andrew.cmu.edu} \\\And
  Gayathri Ganesh Lakshmy \\
  Carnegie Mellon University\\
  \texttt{gganeshl@andrew.cmu.edu} \\\And
  Yiqing Xie \\
  Carnegie Mellon University \\
  \texttt{yiqingxi@andrew.cmu.edu} \\}

\begin{document}

\definecolor{lightgreen}{RGB}{220,240,220}
\definecolor{darkgreen}{RGB}{150,200,150}
\definecolor{lightred}{RGB}{240,220,220}
\definecolor{darkred}{RGB}{200,150,150}
\definecolor{BaselineRed}{RGB}{220, 50, 47}
\definecolor{RerankGreen}{RGB}{150,200,150}
\newcommand\myworries[1]{\textcolor{red}{#1}}

\newcommand\gayathri[1]{\textcolor{purple}{#1}}

\maketitle
\begin{abstract}
Retrieval-Augmented Code Generation (RACG) is a critical technique for enhancing code generation by retrieving relevant information. 
In this work, we conduct an in-depth analysis of code retrieval by systematically masking specific features while preserving code functionality. Our discoveries include: (1) although trained on code, current retrievers heavily rely on surface-level textual features (e.g., docstrings, identifier names), and (2) they exhibit a strong bias towards well-documented code, even if the documentation is irrelevant.
Based on our discoveries, we propose \modelname, a framework that enriches textual information and reduces bias by augmenting code or structural knowledge with semantic information. 
Extensive experiments show that \modelname substantially improves code retrieval (e.g., by 12.8\% / 9.4\% / 7.0\% Recall@1 on HumanEval / MBPP / SWE-Bench-Lite), which also leads to better code generation performance (e.g., by 4.88\% Pass@1 on HumanEval).

\end{abstract}

\section{Introduction}

Retrieval-augmented code generation (RACG) is the technique of generating code based on relevant documents retrieved from a corpus \cite{koziolek2024llm, lu2022reaccretrievalaugmentedcodecompletion}.
RACG is shown to be beneficial in script-level code generation, which provides background knowledge or functionally relevant snippets, and is particularly important for repository-level (repo-level) code generation, where models must be aware of other files within the repository~\cite{wang2025coderagbenchretrievalaugmentcode}.
However, recent work has shown that retrieval quality remains a significant bottleneck for RACG performance (e.g., Agentless \cite{xia2024agentlessdemystifyingllmbasedsoftware} only achieves 35.3\% line localization accuracy on SWE-Bench). 


\begin{figure}[!t]
\includegraphics[width=\linewidth]{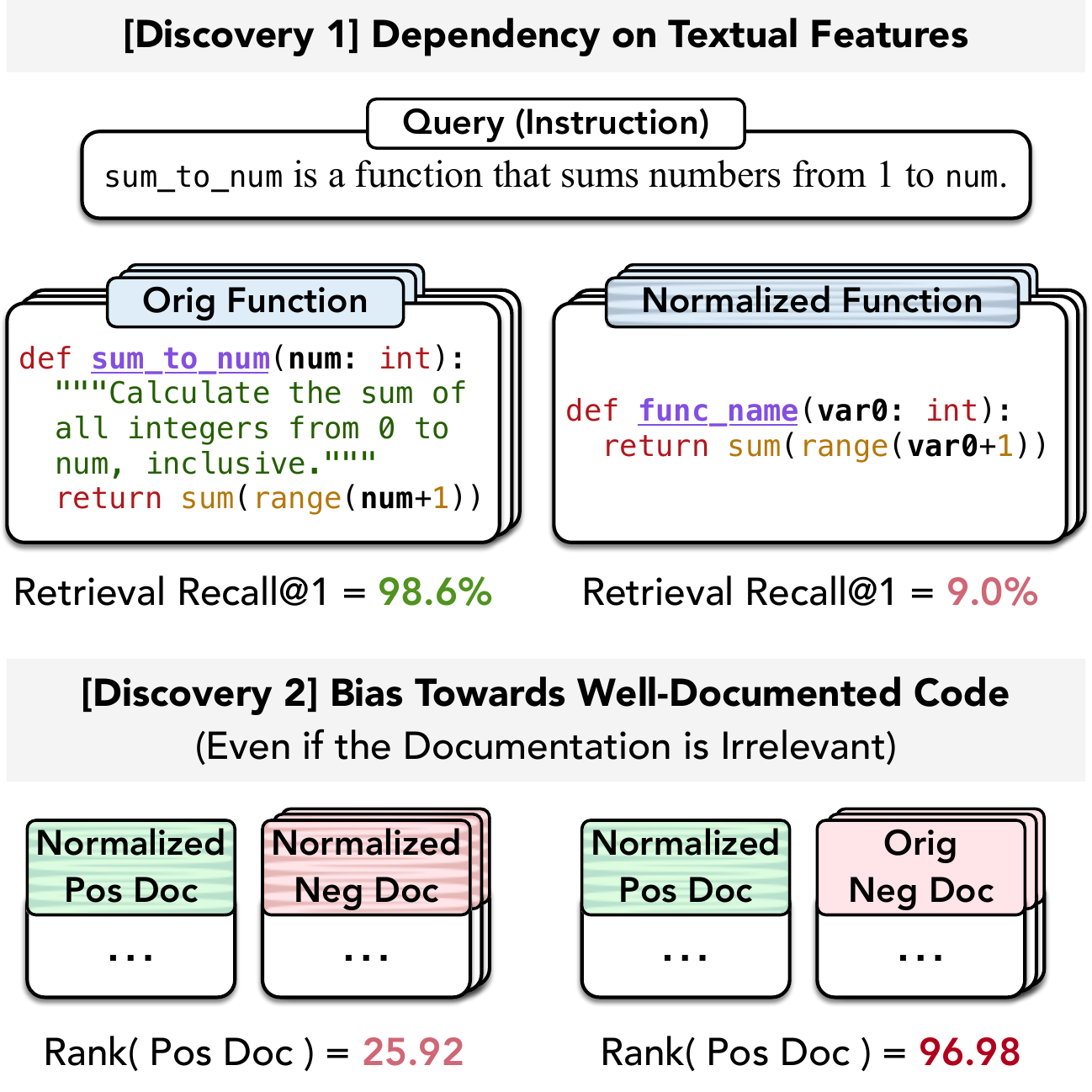}
\centering
\caption{Summaries of our discoveries from the analyses. Our analyses reveal code retrievers' heavy dependence on textual features rather than functional semantics, leading to bias favoring well-documented code regardless of relevance.}
\end{figure}

While extensive analysis has been conducted on the capabilities and challenges of \emph{text retrievers} \cite{Dai_2024, karpukhin2020dense, thakur2021beir}, 
a systematic investigation of \emph{code retrievers} remains relatively underexplored. 
The nature of code corpora differs fundamentally from text corpora due to their highly structured nature, with strict syntax rules and structures \cite{husain2019codesearchnet, allamanis2018survey}.
Unlike text documents, the semantic meaning of a code snippet can be completely altered by small syntactic changes, making traditional retrieval analysis less effective for code search tasks.
Such fundamental differences suggest that code retrievers may have significantly different behaviors from text retrievers, highlighting the need for more focused analysis.


In this work, to develop a deeper understanding of code retrieval, we conduct empirical analyses on both code retrievers and in-context rerankers to answer two critical research questions: \textbf{(RQ1) What features are code retrievers primarily based on?} and \textbf{(RQ2) Do retrievers exhibit bias?} Specifically, we introduce a normalization-based analysis framework, where we systematically mask textual features such as docstrings, function names, and variable names or replace them with placeholders.
Such transformations preserve the code's functionality but eliminate textual cues, allowing us to evaluate the dependence and bias of textual features.

We illustrate our discoveries in Figure 1. We observe that \textbf{[Discovery 1]} \textbf{although trained on code, current code retrievers exhibit strong dependency on textual features} (e.g., docstrings and function or variable names) and under-utilize the functionality of code. 
Specifically, when all textual features are normalized, we observe significant performance degradation on both embedding-based code retrieval and in-context code reranking.
For instance, with normalization, the Recall@1 performance of GIST-large degrades from 98.6\% to 9.0\% on MBPP~\citep{mbpp}.

As shown in Figure 1, our analyses also reveal that \textbf{[Discovery 2]} \textbf{Retrievers consistently assign higher relevance scores to well-documented code, even when the documentation is functionally irrelevant}. 
Particularly, compared to Discovery 1's setting, where all the code documents are normalized, only normalizing the positive documents leads to even worse retrieval and reranking performances.
The results indicate the bias towards well-documented code with meaningful identifier names, which may lead to preferring irrelevant but well-documented code over relevant but poorly documented code.


Based on these discoveries, we present \modelname, which improves code retrieval with \textbf{S}emantic-\textbf{A}ugmented \textbf{C}ode Reranking and in-context \textbf{L}ocalization. 
Based on our discoveries that retrievers are more sensitive to textual features, in the reranking stage, we first generate textual descriptions for the retrieved code documents, and then aggregate the retrieval scores for the original code documents and textual descriptions for final re-ranking.
Such design bridges the code and text modalities and mitigates the bias between well-documented and sparsely documented code.
Based on the empirical discovery that in-context reranking also exhibits similar textual bias, for repo-level code generation, we further introduce semantic-augmented in-context localization, where we generate supplementary file descriptions for the repository structure to augment the context for file localization.
Empirical results show that such methods are the most effective when the file names do not contain rich semantic information.

Our experimental results demonstrate significant improvements across three public benchmarks: HumanEval~\cite{chen2021evaluating}, MBPP~\cite{mbpp}, and SWE-Bench-Lite~\cite{jimenez2023swe}. 
For instance, \modelname achieves 12.8\%/9.4\% code retrieval Recall@1 gain on HumanEval/MBPP under the full normalization setting and achieves 7.0\% file localization Recall@1 on SWE-Bench-Lite with the Agentless pipeline~\cite{xia2024agentlessdemystifyingllmbasedsoftware}.
Our improvements on code retrieval and localization also leads to performance gain on code generation (e.g., 4.88\% Pass@1 gain on HumanEval and 1.67\% on SWE-Bench-Lite).
These results highlight the effectiveness of our approaches in enhancing the semantic understanding capabilities of code retrievers and mitigating lexical bias.


\section{Analysis: Textual Bias in Code Retrieval}
\label{sec:analysis}
\begin{figure*}[!ht]
    \centering
    \includegraphics[width=0.495\textwidth]{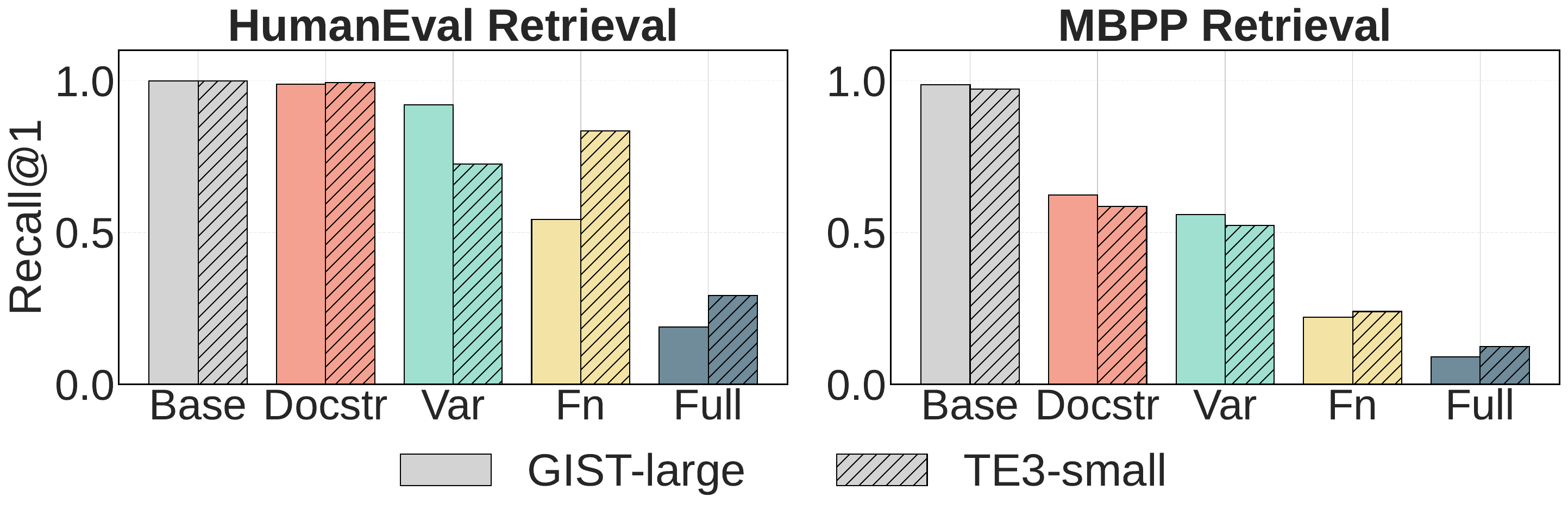}
    \includegraphics[width=0.495\textwidth]{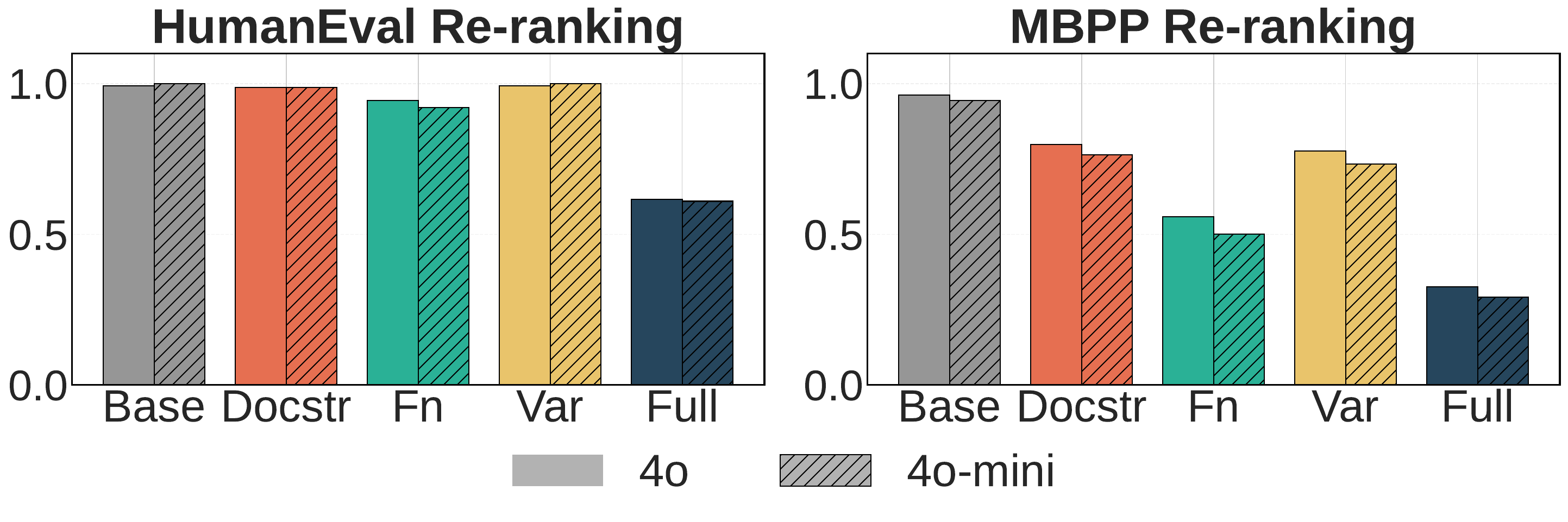}
    \caption{Impact of code normalization techniques on retrieval performance (Recall@1) across datasets with different embedding-based retrievers (\textit{left}) and in-context rerankers (\textit{right}). The results consistently show that all normalization techniques reduce retrieval effectiveness, with function name normalization and full normalization having the most significant negative impact.
    }
    \label{fig:normalization_impact}
\end{figure*}

This section answers our two research questions through two controlled experiments: 
\textbf{(RQ1)} What features are code retrievers primarily based on? and 
\textbf{(RQ2)} Do retrievers exhibit lexical-level bias? 
These experiments aim to identify whether code retrievers favor textual characteristics over functional semantics when matching queries to code.


\subsection{RQ1: What Features are Code Retrievers Based on?}
\label{sec:analysis_feature}

\start{Setup}
We quantify the importance of various features through a controlled study, where we progressively replace surface-level code features with dummy placeholders (e.g., ``func\_0'', ``var\_0'').
We call the process \textbf{``normalization''}.
Specifically, we compare the Recall@1 performance of various code retrievers in five normalization settings:

\circled{1} no normalization (i.e., the original code).
\circled{2} removing docstrings and comments,
\circled{3} renaming function names, along with removing docstrings,
\circled{4} renaming variable names, along with removing docstrings, and
\circled{5} renaming both function names and variable names, along with removing docstrings (i.e., the combination of \circled{2}\textasciitilde \circled{4}).

Note that this study preserves functional equivalence while enabling controlled ablation of specific features with rich textual information.

We study two categories of methods using four models: embedding-based retrievers (GIST-large, TE3-small), which rank the code documents based on the cosine similarity with the query, and LLM-based rerankers (GPT-4o, 4o-mini), where we provide the ground truth (GT) document and the top-50 documents retrieved under the ``no normalization'' setting in the context, and then prompt the LLM to identify the most relevant document.

\start{Results}
As shown in \autoref{fig:normalization_impact}, results on HumanEval and MBPP reveal that both embedding-based retrievers and LLM-based rerankers have significant performance degradation under the normalization settings, especially the full normalization setting. This indicates that they heavily rely on textual features in retrieval or reranking.

We also observe that different models exhibit distinct sensitivities to different normalization methods. For instance, GIST-large shows more severe performance degradation with function name normalization (54.3\% to 18.9\%), while TE3-small suffers greater relative impact when variable names are normalized.
Notably, docstring removal impacts MBPP significantly more than HumanEval (37\% drop in MBPP versus minimal decrease in HumanEval). 
This difference stems from HumanEval's natural language descriptions containing function signatures that exactly match corpus signatures, providing strong retrieval signals even without docstrings, while MBPP's queries have fewer direct lexical matches.

The main discovery of this analysis is that \textbf{[Discovery 1] retrievers heavily rely on textual features, including docstrings and identifier names, rather than deeper semantic information such as the functionality of the code}. One possible explanation is that among the contrastive pairs used for retriever training (e.g., docstring-function pairs and StackOverflow QA pairs)~\cite{husain2019codesearchnet}, the textual queries have a high degree of lexical overlap with docstrings, function names, variable names, etc., and such correlation is captured by the retriever model.

\begin{table}[t]
\centering
\resizebox{\linewidth}{!}{
\begin{tabular}{ccccc}
\toprule

\multicolumn{5}{c}{\textbf{Embedding Model: GIST-large}} \\
\midrule
\multicolumn{3}{c}{Normalization Type} 
& \multirow{2}{*}{\begin{tabular}{c}
(\textbf{S1}) Norm GT \& \\
Norm Others
\end{tabular}}
& \multirow{2}{*}{\begin{tabular}{c}
(\textbf{S2}) Norm GT \& \\
Orig Others
\end{tabular}} \\
Docstring & Var & Func & & \\
\midrule
\xmark & \xmark & \xmark & 1.00 & 1.00 \\
\cmark & \xmark & \xmark & 1.01 & \bad 1.07 \\
\cmark & \cmark & \xmark & 1.35 & \bad 5.18 \\
\cmark & \xmark & \cmark & 8.05 & \worse 20.29 \\
\cmark & \cmark & \cmark & 87.18 & \worst 288.27 \\
\midrule
\multicolumn{5}{c}{\textbf{Embedding Model: OpenAI/text-embedding-3-small}} \\
\midrule
\multicolumn{3}{c}{Normalization Type} 
& \multirow{2}{*}{\begin{tabular}{c}
(\textbf{S1}) Norm GT \& \\
Norm Others
\end{tabular}}
& \multirow{2}{*}{\begin{tabular}{c}
(\textbf{S2}) Norm GT \& \\
Orig Others
\end{tabular}} \\
Docstring & Var & Func & & \\
\midrule
\xmark & \xmark & \xmark & 1.00 & 1.00 \\
\cmark & \xmark & \xmark & 1.01 & \bad 1.04 \\
\cmark & \cmark & \xmark & 2.23 & \worse 22.48 \\
\cmark & \xmark & \cmark & 1.74 & \bad 1.95 \\
\cmark & \cmark & \cmark & 25.92 & \worst 96.98 \\
\bottomrule
\end{tabular}
}
\caption{Average Rank of the GT document ($\downarrow$) on the HumanEval dataset. The results demonstrate that when only the ground truth document is normalized (S2) while others remain in their original form, the GT document's rank deteriorates dramatically compared to when all documents are normalized (S1). Such results reveal a strong bias toward textual features over semantic relevance.
}
\label{tab:avg_rank}
\end{table}

\subsection{RQ2: Do retrievers exhibit bias?}
\label{sec:analysis_bias}

\start{Setup}
Following \S\ref{sec:analysis_feature}, we further investigate whether code retrievers have a bias towards code containing more or fewer textual features (e.g., docstrings and function/variable names).
Towards this goal, unlike the setting in \S\ref{sec:analysis_feature} (\textbf{S1}), where all the documents are normalized,
we introduce an asymmetric normalization setting (\textbf{S2}), where only the ground truth (GT) document for each query is normalized, while the remainder of the corpus is left in its original form.
The comparison of the retrievers' performances under S1 and S2 allows us to assess whether models penalize stylistic deviations in semantically equivalent code.

\start{Results}
As shown in \autoref{tab:avg_rank}, the retrievers' performance further decreases when the irrelevant code documents are more well-documented than the ground truth one.
For instance, in the most extreme case where all the identifiers and docstrings are normalized, the rank of the GT document jumps from 87.18 to 288.27 for GIST-large retrieval (and from 25.92 to 96.98 for TE3-small).

This degradation reveals a clear inductive bias in current retrieval models: in many cases, the retriever assigns a higher rank to irrelevant but well-documented code over the semantically correct, normalized gold document.
In other words, we observe that \textbf{[Discovery 2] retrievers tend to assign higher scores for well-documented code with meaningful identifier names, even if the documentation is irrelevant to the query}.



\section{Methodology}


Both \textbf{[Discovery 1]} and \textbf{[Discovery 2]} reveal that code retrievers heavily rely on textual information (e.g., documentation and identifier names) rather than understanding of code structure.
To combat this issue, \modelname introduces two techniques: semantic-augmented code reranking and semantic-augmented in-context localization.
Both methods augment the retrieved code or structure with textual descriptions to improve the encapsulation of semantic information.

\subsection{Semantic-Augmented Code Reranking}

Traditional retrieval systems often struggle with the semantic gap between natural language queries and code documents. 

To bridge this gap, we enhance the re-ranking process with semantically rich descriptions. After retrieving the initial top-$k$ code documents, we prompt an LLM to generate concise natural language descriptions of each code snippet's functionality and purpose. These descriptions provide an alternative representation of the code that emphasizes semantic content over syntactic structure.

Then we combine the relevance scores between (1) the original code documents and the queries, and (2) the textual descriptions and the queries:

\begin{equation}
    Score_{final} = (1-\alpha) \cdot Score_{code} + \alpha \cdot Score_{desc}
\label{eq:reranking}
\end{equation}

where $\alpha \in [0,1]$ is a tunable hyperparameter controlling the influence of each score component.

This approach transforms the cross-modal comparison problem (text-to-code) into a more tractable text-to-text comparison, enabling more semantically meaningful ranking of code documents based on natural language queries.

\begin{figure}[t]
\includegraphics[width=\linewidth]{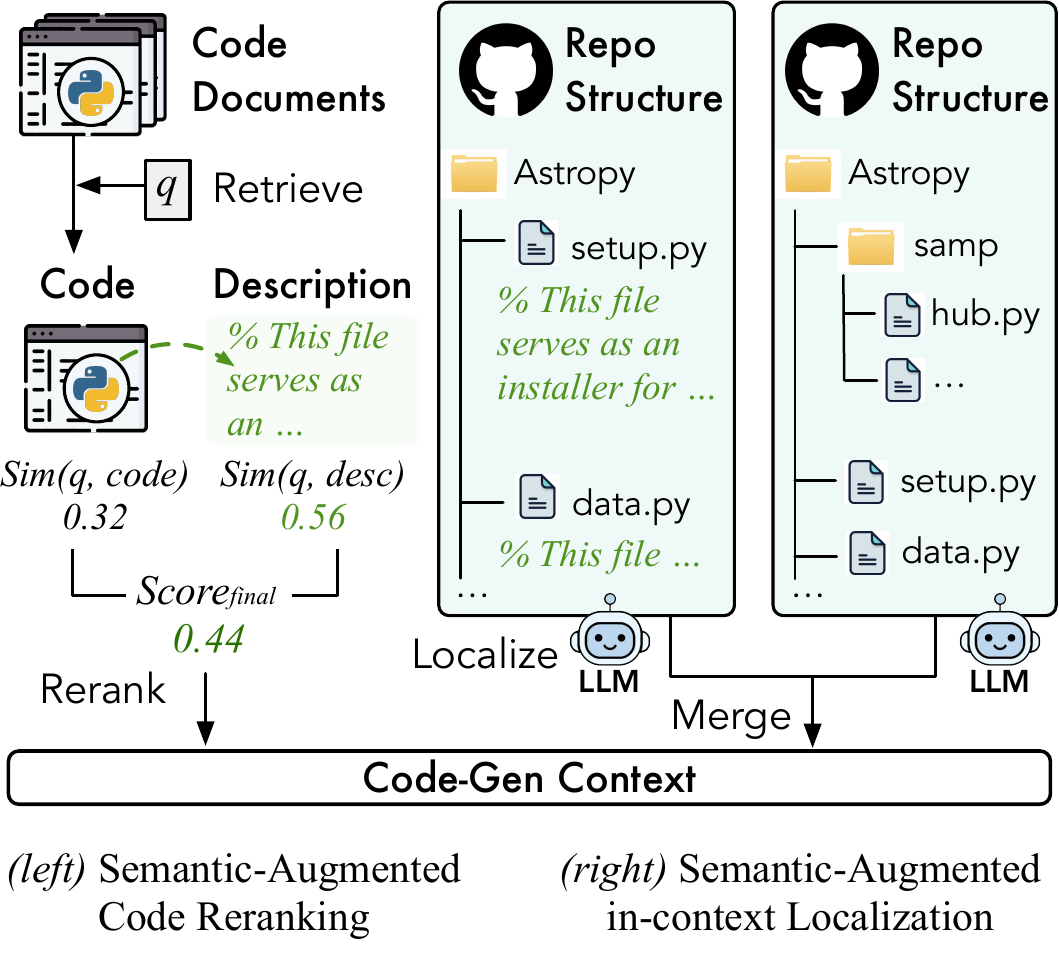}
\centering
\caption{ Flowchart illustrating our two main approaches. \textit{(left)} Documents are retrieved based on query similarity, then augmented with generated descriptions. The final ranking score combines both code-query and description-query similarity scores to improve retrieval performance. \textit{(right)} Repository structure is enhanced with descriptive summaries for each file (shown in green italics). This augmented structure significantly improves the LLM's ability to localize relevant files for code modification tasks.}
\label{fig:repo_structure}
\end{figure}

\subsection{Semantic-Augmented in-context Localization}




Repository-level code generation required comprehensive understanding of code repository structure to effectively navigate and modify complex codebases \cite{xia2024agentlessdemystifyingllmbasedsoftware}. When previous works such as \cite{xia2024agentlessdemystifyingllmbasedsoftware} uses repository structures to localize code that needs editing, they present the structure in a hierarchical format that represents the directory and file organization, as illustrated in Figure \ref{fig:repo_structure}. 

Building on our discoveries that retrievers tend to focus on textual information rather than functional semantics, we aim to augment structural representations with rich textual descriptions to improve localization performance. 
We enhance the standard repository structure by generating brief semantic descriptions for each file's contents in the repository. The descriptions are generated by prompting an LLM to analyze each file individually and summarize its contents (classes, functions, etc.), purpose and functionality in a couple sentences. 
The augmented repository structure serves as input to an LLM tasked with identifying potentially suspicious files related to a reported bug or issue. These semantic descriptions allow the model to better understand file functionality and relationships, improving subsequent localization steps by reducing noise. 
To improve efficiency in inference, we generate descriptions using a small (Llama-3.1-8B-based) file summarization model, which already shows significant performance gain in file localization.

\section{Experiments}

\begin{table*}[ht]
\centering
\resizebox{\linewidth}{!}{
\begin{tabular}{lccccccccc|ccccccccc}
\toprule
\multirow{3}{*}{\begin{tabular}{l}
\bf Norm. \\ \bf Type
\end{tabular}}
& \multicolumn{9}{c|}{\textbf{HumanEval}} 
& \multicolumn{9}{c}{\textbf{MBPP}} \\
& \multicolumn{3}{c}{\bf Recall@1} 
& \multicolumn{3}{c}{\bf Recall@5} 
& \multicolumn{3}{c|}{\bf Recall@10} 
& \multicolumn{3}{c}{\bf Recall@1} 
& \multicolumn{3}{c}{\bf Recall@5} 
& \multicolumn{3}{c}{\bf Recall@10} \\
\cmidrule(lr){2-4} \cmidrule(lr){5-7} \cmidrule(lr){8-10}
\cmidrule(lr){11-13} \cmidrule(lr){14-16} \cmidrule(lr){17-19}
 & Base & \modelname & $\Delta$ 
 & Base & \modelname & $\Delta$ 
 & Base & \modelname & $\Delta$
 & Base & \modelname & $\Delta$ 
 & Base & \modelname & $\Delta$ 
 & Base & \modelname & $\Delta$ \\
\midrule
Docstring & 98.8 & 98.8 & 0.0 
& 100.0 & \better 100.0 & \better 0.0 
& 100.0 & \better 100.0 & \better 0.0 
& 62.4 & \better 70.2 & \better $\uparrow$7.8 
& 87.0 & \good 89.2 & \good $\uparrow$2.2 
& 91.4 & \good 93.2 & \good $\uparrow$1.8 \\
Func Name & 54.3 & \better 69.5 & \better $\uparrow$15.2 
& 77.4 & \better 86.0 & \better $\uparrow$8.6 
& 83.5 & \better 89.6 & \better $\uparrow$6.1
& 22.0 & \better 36.8 & \better $\uparrow$14.8 
& 39.8 & \better 51.0 & \better $\uparrow$11.2 
& 49.0 & \better 55.0 & \better $\uparrow$6.0 \\
All & 18.9 & \better 31.7 & \better $\uparrow$12.8 
& 34.1 & \better 43.3 & \better $\uparrow$9.2 
& 42.7 & \good 46.3 & \good $\uparrow$3.6
& 9.0 & \better 18.4 & \better $\uparrow$9.4 
& 18.6 & \better 29.0 & \better $\uparrow$10.4 
& 23.6 & \better 31.4 & \better $\uparrow$7.8 \\

\bottomrule
\end{tabular}}
\caption{The Recall@$k$ retrieval performance of the baseline and \modelname under different normalization settings. We use GIST-large as the retriever and Llama-3.1-8B-Instruct for generating descriptions. We highlight results showing \textit{\modelname > Base} with green (darker green when having 5\%+ increases or perfect results).
}
\label{tab:script_retrieval}
\end{table*}

\begin{table}[t]
\centering
\resizebox{0.85\linewidth}{!}{
\begin{tabular}{lcccc}
\toprule
\multirow{2}{*}{\bf Method} 
& \multicolumn{2}{c}{\bf Localization Accuracy} \\
& Line & File ($\Delta$) \\
\midrule
Agentless (GPT-4o-mini) & 32.7 & 70.0 \\
\quad + \modelname & \good 34.7 ($\uparrow$2.0) & \better 78.0 ($\uparrow$8.0) \\
Agentless (GPT-4o) & 40.0 & 79.0 \\
\quad + \modelname & \good 42.3 ($\uparrow$2.3) & \better 86.0 ($\uparrow$7.0) \\
\bottomrule
\end{tabular}
}
\caption{Fault Localization results on SWE-Bench-Lite. We compute the \% of instances where the retrieved/LLM-localized files/lines contain the fault location. File-level localization is computed using the list of potential files identified by agentless at the end of the localization phase. Line-level localization is computed using final patches after testing and re-ranking.}
\label{tab:agentless-retrieval}
\end{table}

\begin{table*}[h]
\centering
\resizebox{0.85\linewidth}{!}{
\begin{tabular}{lcccccccccccc}
\toprule
\multirow{3}{*}{\begin{tabular}{l}
\bf Normalization \\ \bf Type
\end{tabular}}
& \multicolumn{6}{c}{\textbf{HumanEval Pass@1}} 
& \multicolumn{6}{c}{\textbf{MBPP Pass@1}} \\
& \multicolumn{3}{c}{\bf Qwen2.5-Coder-7B} 
& \multicolumn{3}{c}{\bf Deepseek-coder-7b} 
& \multicolumn{3}{c}{\bf Qwen2.5-Coder-7B} 
& \multicolumn{3}{c}{\bf Deepseek-coder-7b}  \\
\cmidrule(lr){2-4} \cmidrule(lr){5-7}
\cmidrule(lr){8-10} \cmidrule(lr){11-13}
& Base & \modelname & $\Delta$
& Base & \modelname & $\Delta$
& Base & \modelname & $\Delta$
& Base & \modelname & $\Delta$ \\
\midrule
Docstring & 99.39 & 99.39 &  0.00
& 99.39 & 99.39 & 0.00
& 57.40 & \better 61.60 & \better $\uparrow$4.20
& 57.40 & \better 61.60 & \better $\uparrow$4.20
\\
Func Name & 99.39 & 99.39 & 0.00  
& 99.39 & 99.39 & 0.00 
& 19.00 & \better 24.40 & \better $\uparrow$5.40
& 19.00 & \better 24.40 & \better $\uparrow$5.40
\\
All & 93.29 & \better 98.17 & \better $\uparrow$4.88
& 93.29 & \better 98.17 & \better $\uparrow$4.88
& 6.80 & \good 9.00 & \good $\uparrow$2.20
& 6.80 & \good 9.00 & \good $\uparrow$2.20
\\
\bottomrule
\end{tabular}}
\caption{The Pass@1 performance of the baseline and \modelname on HumanEval and MBPP.}
\label{tab:script_generation}
\end{table*}

\begin{table}[t]
\centering
\resizebox{\linewidth}{!}{
\begin{tabular}{lcc}
\toprule
\bf Method
& \bf \% Non-Empty ($\Delta$) 
& \bf \% Resolved ($\Delta$) \\
\midrule
Agentless (GPT-4o-mini) & 93.3 & 14.7 \\
\quad + \modelname & \good 94.3 ($\uparrow$1.0) & \good 16.0 ($\uparrow$1.3) \\
Agentless (GPT-4o) & 97.0 & 24.3 \\
\quad + \modelname & \good 98.0 ($\uparrow$1.0) & \good 26.0 ($\uparrow$1.7) \\
\bottomrule
\end{tabular}
}
\caption{Code generation results on SWE-Bench-Lite.}
\label{tab:agentless-generation}
\end{table}



With our experiments, we aim to answer the following research questions: \textbf{(RQ1)} What impact does \modelname have on Code retrieval performance? \textbf{(RQ2)} Given this code retrieval performance, what is the downstream code generation performance improvement? \textbf{(RQ3)} Why does semantic augmentation benefit code retrieval? \textbf{(RQ4)} Which hyper-parameters are optimal? 

\subsection{Experiment Setup}
\start{Datasets} 
We evaluate our approach on three widely used benchmarks:
HumanEval \cite{chen2021evaluating} and MBPP \cite{mbpp} are script-level algorithm problem datasets and SWE-Bench-Lite~\cite{jimenez2023swe} is a repo-level issue-solving dataset.
For HumanEval and MBPP, we evaluate under various ``normalization'' settings (as introduced in \S\ref{sec:analysis}), which preserve the functionalities of the code documents but are more challenging to retriever models.

\start{Evaluation Metrics}
We follow existing work~\cite{wang2025coderagbenchretrievalaugmentcode} and report Recall@$k$ ($k$=1,5,10) for code retrieval and report Pass@1 for code generation.
We additionally evaluate file and line localization accuracy~\cite{xia2024agentlessdemystifyingllmbasedsoftware} for SWE-Bench-Lite, which checks whether the corresponding generated patch edits a superset of all locations in the ground truth patch.

\start{Implementation Details}
For scalability concerns, we only generate a short description (under 100 words) for the code documents using a small model (Llama-3.1-8B-Instruct).
For SWE-Bench Lite, we integrate our approach into the file localization step of the Agentless pipeline \cite{xia2024agentlessdemystifyingllmbasedsoftware}, which prompts an LLM to identify relevant files based on the repository structure in the format of a tree.

We provide more experimental details in \ref{sec:implementation_details}.


\subsection{Code Retrieval Results}
\start{Script-level Code Generation Results}
\autoref{tab:script_retrieval} presents code retrieval results under HumanEval and MBPP across different normalization settings, helping us answer \textbf{(RQ1)}.
We observe that \modelname demonstrates significant improvements over the baseline (e.g., improving Recall@1 for up to 15.2\% on HumanEval and 14.8\% on MBPP).
Particularly, under the most challenging setting where all textual features are normalized, \modelname still obtains substantial performance gain, which indicates that LLMs can still effectively summarize the functionalities even if all textual features are normalized. Our approach leverages LLM's strong code understanding property to compensate code retriever's bias toward textual features and hence effectively capture the semantic meaning of code.
 

\start{Repo-level Issue-Solving Results}
As shown in \autoref{tab:agentless-retrieval}, \modelname achieves significant performance gain on fault localization on SWE-Bench-Lite. For instance, we improve the file localization accuracy by 8.0\%/7.0\% for 4o-mini/GPT-4o.
The consistent performance gain across different models highlights the effectiveness of augmenting repository structures with richer contextual information for fault localization.
Specifically, as shown in later analysis (\S\ref{sec:exp_analysis}), the file descriptions may contain high-level descriptions of the file's purpose, its relationship to other files, or its utility to the whole repository, which are neglected in the repository tree structure.
Such information reveals the high-level role and interconnections of files in the repository, which are relevant to the issues.

\subsection{Code Generation Results}
\begin{figure}[t]
    \centering
    \vspace{-0.1cm}
    \includegraphics[width=\linewidth]{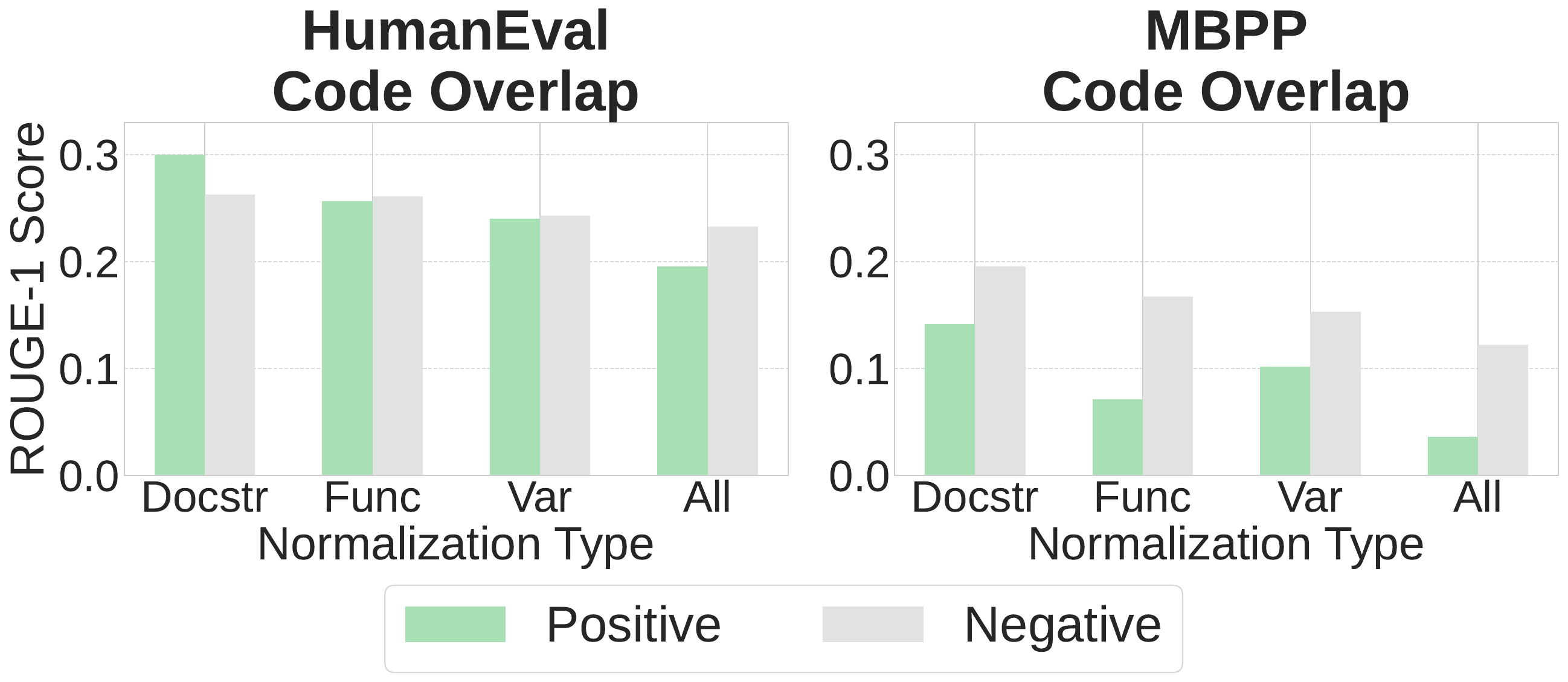} \\
    \includegraphics[width=\linewidth]{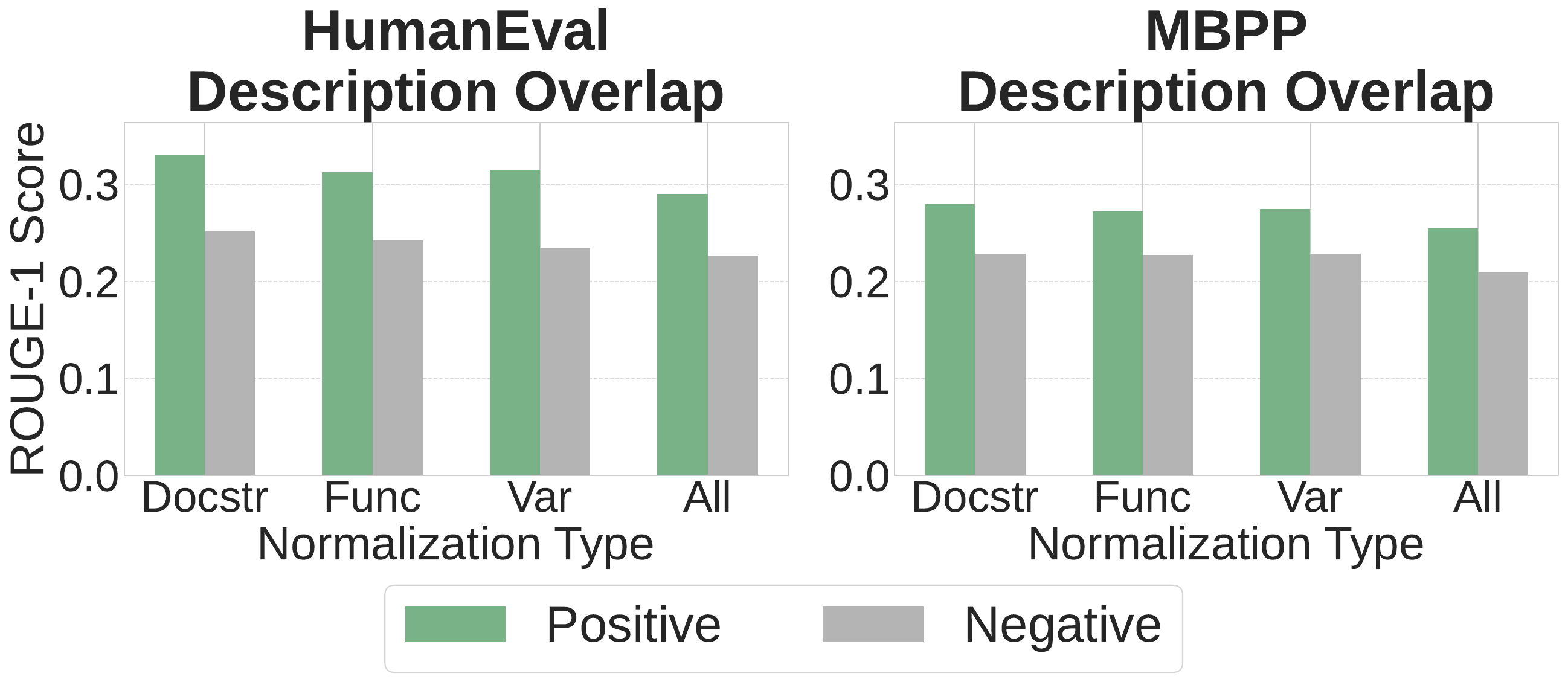}
    \caption{Lexical overlap (ROUGE-1 scores) between the query and the positive/negative code/descriptions in HumanEval and MBPP. 
    We show the negative example with the highest overlap with the query.
    While under the normalization conditions, the negative code has similar or even higher lexical overlap with the query than the positive one, the positive descriptions always have higher ROUGE-1 scores.
    }
    \label{fig:lexical_overlap}
\end{figure}

\start{Script-level Code Generation Results}
Results in \autoref{tab:script_generation} demonstrate that our performance gain in retrieval also translates to the improvement in code generation for various code generation models, highlighting the robustness of our method, tackling our \textbf{(RQ2)}. 
On MBPP, which presents a more challenging scenario due to lower lexical overlap between queries and relevant code (as shown in \autoref{fig:lexical_overlap}), we observe improvements across all normalization settings.
Note that under the most challenging setting of full normalization, where even providing the normalized GT documents in the context only gives marginal code generation performance gain, \modelname still delivers an improvement of 2.2 Pass@1.
These results demonstrate that better retrieval directly translates to improved generation performance, with the benefits being most pronounced in scenarios where code lacks rich textual features.

\begin{table}[t]
\centering
\small
\begin{tabular}{lcc}
\toprule
\textbf{Category} & \textbf{Frequency (\%)} & \textbf{Gain (\%)} \\
\midrule
Functional Purpose & 300 (100.00\%) & \better $\uparrow$24 (8.00\%) \\
Core Components & 242 (80.67\%) & \better $\uparrow$20 (8.26\%) \\
File Relations & 42 (14.00\%) & \better $\uparrow$4 (9.52\%) \\
\bottomrule
\end{tabular}
\caption{Analysis of the file descriptions used by \modelname. We manually design the categories based on the descriptions' content and use 4o-mini for categorization.}
\label{tab:description-categories}
\end{table}

\begin{figure}[t]
    \centering
    \includegraphics[width=\linewidth]{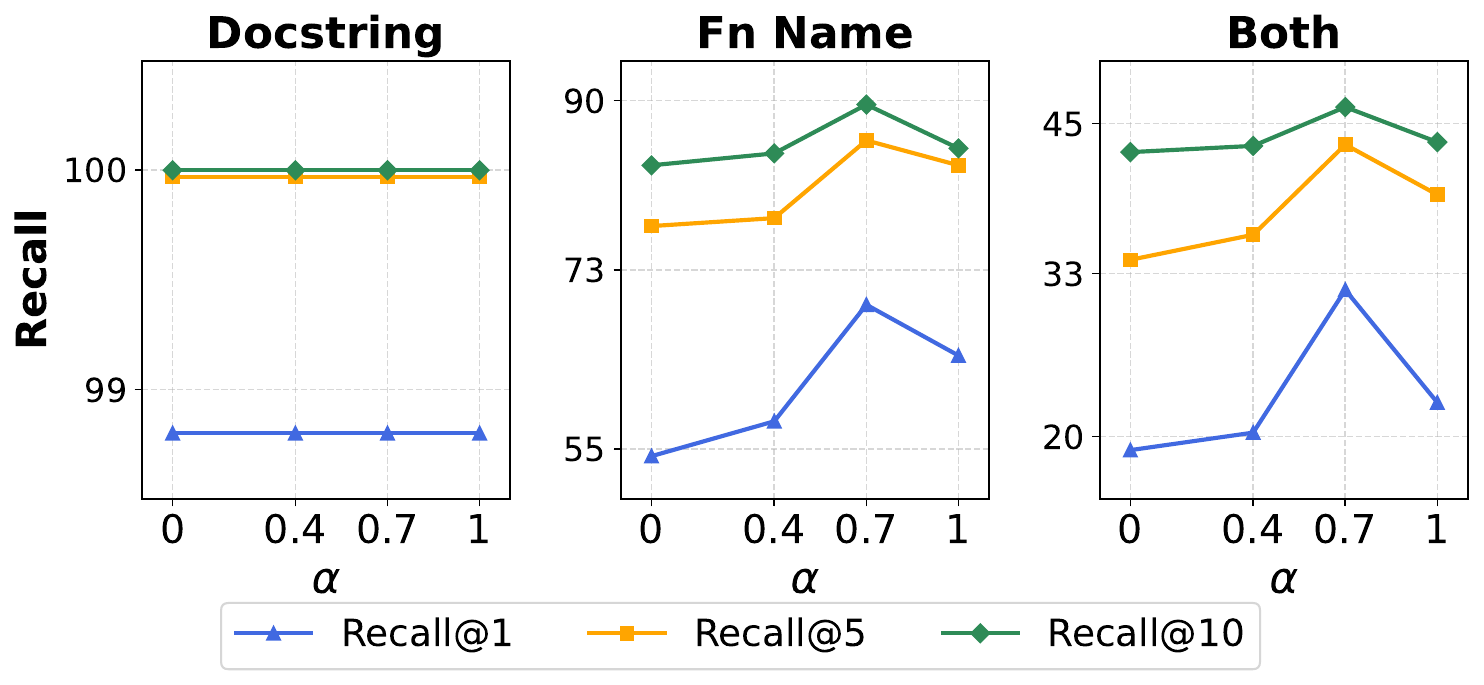}
    \caption{Impact of normalization parameter $\alpha$ on Recall performance. The three plots show performance across different normalization approaches: Docstring (left), Function Name (middle), and Both - Functions and Variables (right).}
    \vspace{-0.2cm}
    \label{fig:hyperparameter_analysis}
\end{figure}

\start{Repo-level Issue-Solving Results} 
As shown in \autoref{tab:agentless-generation}, our semantic-augmented approach also improves issue-solving rates on SWE-Bench-Lite (e.g., by 1.7\% Pass@1 over Agentless (GPT-4o)). 
This is consistent with previous work's observation that fault localization accuracy is a bottleneck for repo-level code generation~\cite{xia2024agentlessdemystifyingllmbasedsoftware}.
The improvement in resolved issues is particularly crucial as it further validates our approach's effectiveness in practical software engineering scenarios.

\begin{figure*}[t]
\includegraphics[width=0.9\linewidth]{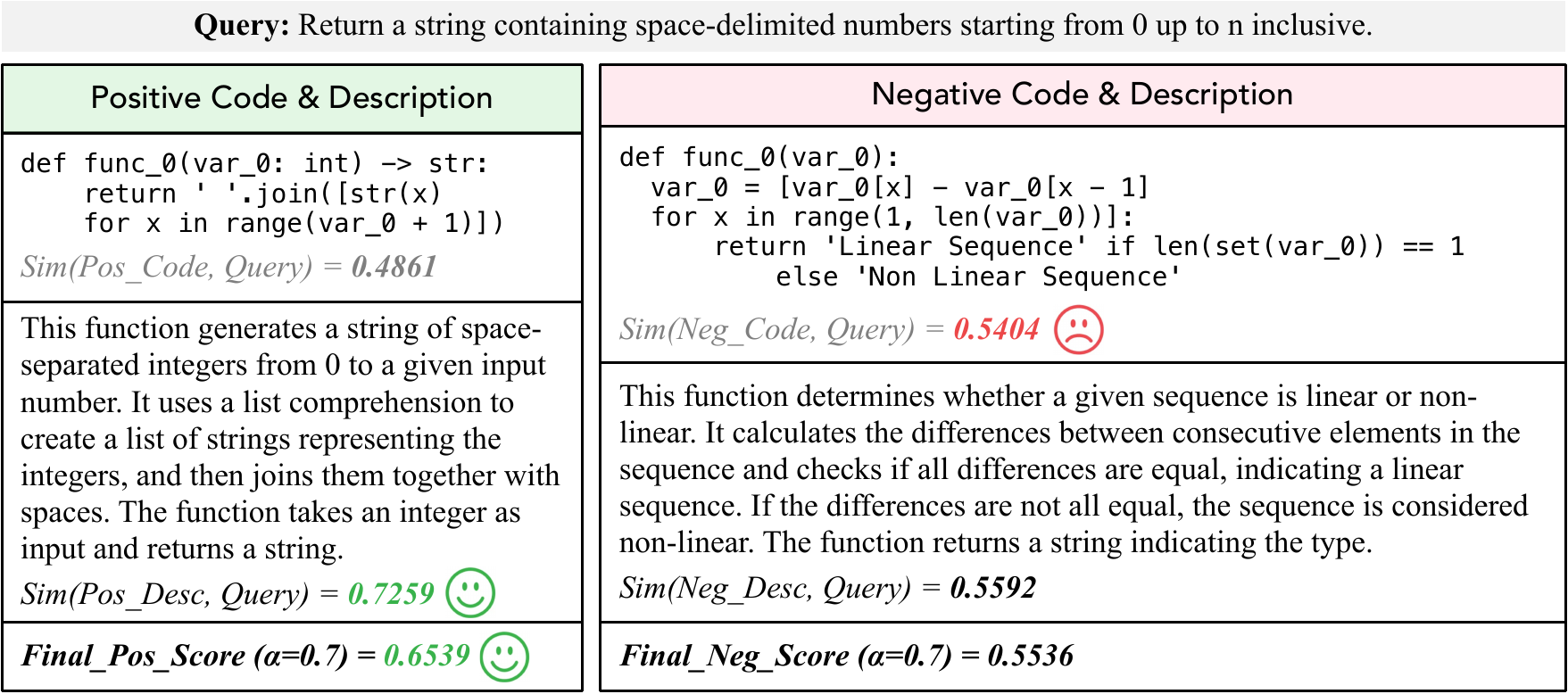}
\centering
\caption{Case Study 1 (HumanEval/15): The baseline incorrectly ranks a function that checks for linear sequences higher based on code similarity alone, but semantic-augmented reranking correctly identifies the ground truth solution by leveraging description similarity.}
\label{tab:case_study_1}
\end{figure*}

\subsection{Performance Analysis}
\label{sec:exp_analysis}

\start{\modelname Improves Query-Doc Lexical Overlap}
To investigate \textbf{(RQ3)}, we hypothesize that semantic descriptions enhance code retrievers by improving lexical overlap between the query and documents, especially when the surface-level features in code are normalized. 
To test our hypothesis, we compare the ROUGE scores between the query and the positive and negative examples.

Results in \autoref{fig:lexical_overlap} reveal that the lexical overlap between positive and negative descriptions is larger than that of code documents.
Particularly, on the challenging MBPP dataset, the positive document's lexical overlap with the query is on average lower than that of the best negative document under the normalization settings, while descriptions maintain a clear separation between positive and negative examples even under full normalization.

\start{\modelname Enriches Context with File Semantics} 
To help understand why our in-context localization method is effective for repo-level code generation, we analyze the contents of generated descriptions for the GT files in SWE-Bench-Lite. Particularly, we define three categories for the descriptions' content: (1) Functional Purpose - the file's overall functionality; (2) Core Components - specific functions, classes, or data structures; and (3) File Relationships - connections to other repository files. Then we use GPT-4o-mini to categorize each description of the GT files.

Answering \textbf{(RQ3)}, our analysis shows that file descriptions serve multiple purposes simultaneously. As shown in \autoref{tab:description-categories} all descriptions (100\%) cover Functional Purpose, 80.67\% describe Core Components, and 14.00\% mention File Relationships. 
This semantic enrichment contributes to the overall 8\% gain in file-level localization, demonstrating how augmented context helps models better understand file purposes and interrelationships within repositories. 
Within the core components and file relations categories, we observe a respective 8.26\% and 9.52\% gain. Both are slightly higher than the overall gain. 

\start{Hyper-Parameter Analysis}
To answer \textbf{(RQ4)}, we illustrate the impact of the weighting parameter $\alpha$ across different normalization approaches.
As shown in \autoref{fig:hyperparameter_analysis}, for all Recall@$k$ values ($k$=1, 5, 10), the performance consistently peaks at $\alpha$=0.7, except in the case of docstring normalization, where all methods perform nearly perfectly. In contrast, pure-description retrieval ($\alpha$=1) and pure-code retrieval (the baseline, with $\alpha$=0) both show lower performance. 
This result indicates that \modelname achieves a balance between retrieval based on code snippets and descriptions, where code preserves full information and descriptions capture the semantics that may be challenging to encode from the code.


\subsection{Case Studies}
We present two representative case studies to illustrate how \modelname effectively bridges the semantic gap between natural language queries and code, especially under normalization settings.


\start{Semantic-Augmented Reranking Example}
As shown in \autoref{tab:case_study_1}, for a query asking to generate space-separated integers, the baseline mistakenly prioritizes an unrelated function analyzing sequence linearity. In contrast, when reranking based on generated descriptions, the correct function receives a higher relevance score. The combined score in \modelname still correctly prioritizes the relevant code, lifting it from rank 18 to the top. This improvement arises because the descriptions capture high-level functional intent—e.g., “generates a string of integers”—which is lost when identifiers and docstrings are normalized. The reranker can thus discriminate relevance based on semantic meaning rather than textual overlap, correcting the baseline’s lexical bias.



\start{In-Context Localization Example} 
Similarly, \autoref{tab:case_studies_loc} presents two cases where our semantic-augmented localization method correctly identifies the source files for real-world GitHub issues but the Agentless baseline does not. For instance, in the first issue, the baseline misattributes the issue to the field accessor logic in ('\_\_init\_\_.py').
In comparison, with the file description that mentions the ``Choices'' class, \modelname locates the correct file, ``enums.py'', where the faulty ``enum'' string representation is actually defined.
Similarly, for the second issue, our method accurately links the issue to the sign function in complexes.py, rather than applying a generic patch to the core function class. 
We hypothesize that these successes stem from our file-level descriptions, which often include module-level purposes and inter-file relationships—semantic signals absent in raw filenames or directory trees.

\section{Related Work}

\start{Code Retrieval and Retrieval-Augmented Code Generation}
Retrieval-augmented code generation (RACG) incorporates external retrieved snippets into the generation pipeline to improve performance. Recent works such as ReACC \cite{lu2022reaccretrievalaugmentedcodecompletion} and Repocoder \cite{zhang2023repocoderrepositorylevelcodecompletion} demonstrate gains by supplying functionally relevant examples during generation. Similarly, Code-RAG Bench \cite{wang2025coderagbenchretrievalaugmentcode} offers a standardized benchmark for evaluating RACG systems across programming tasks, retrieval quality, and computational efficiency. Other efforts extend RACG to novel applications such as universal information extraction to generate task-specific-extractors \cite{guo2023retrievalaugmentedcodegenerationuniversal}.
While these methods highlight the promise of RACG, they implicitly assume effective retrievers, which is not practical.
Our work contributes to this area by revealing and combating a core limitation in current retrievers: a strong bias toward superficial lexical signals. 

\start{LLM-Based Fault Localization}
Recent work has shifted towards complex, real-world scenarios through benchmarks like SWE-Bench \cite{jimenez2023swe}, featuring actual GitHub issues requiring codebase comprehension and bug-fixing.
In parallel, LLM capabilities have enabled significant advances in fault localization (FL). FlexFL \cite{10934742} incorporates open-source LLMs in a two-stage process to leverage bug-related information to identify and refine buggy locations. AgentFL \cite{qin2025agentflscalingllmbasedfault} models FL as a human-like process with specialized agents for comprehension, navigation, and confirmation steps. Agentless \cite{xia2024agentlessdemystifyingllmbasedsoftware} utilize a simplistic three-phase approach to narrow down fault locations from file-level locations to line-level. 
Unlike previous approaches focused on improving agent architectures or specialized tools, \modelname tackles a fundamental limitation in code retrievers themselves: their reliance on surface-level textual features rather than functional semantics.

\vspace{0.25cm}
\section{Conclusion}

We conduct systematic normalization experiments and uncover two key biases for current code retrieval systems: (1) heavy dependence on textual cues over semantic understanding, and (2) consistent preference for well-documented code, even if the documentation is irrelevant. To address these issues, we propose \modelname, a framework that augments code retrieval and localization with semantic information through natural language descriptions.
Experiments demonstrate \modelname's effectiveness in both code retrieval and code generation on three widely used benchmarks: HumanEval, MBPP, and SWE-Bench-Lite. For instance, \modelname improves code retrieval by up to 12.8\% Recall@1 on HumanEval and up to 8.0\% file localization accuracy on SWE-Bench-Lite. 
These gains translate directly into better downstream code generation quality (e.g., up to 4.88\% Pass@1 gain on HumanEval).


\newpage
\section*{Limitations}

Some of our method's limitations are: (1) We do not  attempt to apply our semantic augmentation techniques to agentic methods for code generation, (2) We do not explore the performance of \modelname on other Repo-level coding benchmarks, Lastly, (3) We focus primarily on function retrieval when analyzing lexical-level bias in current code retrieval techniques.


\newpage
\appendix
\section{Appendix}

\subsection{Experimental Details}
\label{sec:implementation_details}

\start{Datasets}
We evaluate our approach on three widely-used code benchmarks: Humaneval \cite{chen2021evaluating} consisting of 164 hand-written programming problems with function signatures and test cases; MBPP \cite{mbpp} containing 974 Python programming tasks with natural language descriptions and test cases; and SWE-Bench Lite \cite{jimenez2023swe} with 300 real-world  GitHub tasks representing realistic software engineering scenarios.

\start{Implementation Details}
For our code reranking approach, we use GIST-large \cite{solatorio2024gistembed} as the base embedding model for code retrieval. We employ Llama-3.1-8B-Instruct to generate concise descriptions (under 100 words) for each code snippet, focusing on functionality, algorithm, and purpose. The prompt template instructs the model to analyze the code and provide a clear description of what it does. 
For reranking, we combine the embedding similarity scores from both code and descriptions using \autoref{eq:reranking} with $\alpha=0.7$. 

As for code generation experiments, for HumanEval for MBPP, we experiment with state-of-the-art coding models: Qwen2.5-Coder-7B \cite{hui2024qwen2} and Deepseek-coder-7b \cite{guo2024deepseekcoderlargelanguagemodel}. 
For SWE-Bench Lite, we integrate our approach with the Agentless pipeline \cite{xia2024agentlessdemystifyingllmbasedsoftware}, using both GPT-4o-mini and GPT-4o for both issue solving and the file summarization step in \modelname.
More specifically, we augment the localization steps for relevant and irrelevant file identification.

\begin{table*}[t]
\centering
\begin{tabular}{|p{0.46\textwidth}|p{0.46\textwidth}|}
\hline
\multicolumn{2}{|p{\textwidth}|}{
\vspace{1mm}
\textbf{Query:} Given an array with non-negative integer nodes, pluck the smallest even integer. If multiple found, return the smallest. If none found, return [].
\vspace{1mm}
} \\
\hline
\multicolumn{1}{|c|}{\textbf{Positive Code}} & \multicolumn{1}{c|}{\textbf{Positive Code Description}} \\
\hline
\begin{minipage}[t]{\linewidth}
\vspace{2mm}
\begin{verbatim}
def func_0(var_0):
    if len(var_0) == 0:
        return []
    var_1 = list(filter(lambda x: 
                x % 2 == 0, var_0))
    if var_1 == []:
        return []
    return [min(var_1), 
            var_0.index(min(var_1))]
\end{verbatim}
\vspace{2mm}
\centering \textcolor{BaselineRed}{\textbf{Code Score:} {0.4713}}
\vspace{2mm}
\end{minipage} & 
\begin{minipage}[t]{\linewidth}
\vspace{2mm}
This function takes a list of integers as input and returns a list containing the smallest even number in the input list and its index. If the input list is empty or contains no even numbers, the function returns an empty list. The function uses list comprehension and the built-in filter function to find even numbers, and then finds the minimum and its index using the min and index methods.

\vspace{3mm}
\centering \textcolor{RerankGreen}{\textbf{Description Score:} 0.6383}
\vspace{2mm}
\end{minipage} \\
\hline
\multicolumn{1}{|c|}{\textbf{Negative Code}} & \multicolumn{1}{c|}{\textbf{Negative Code Description}} \\
\hline
\begin{minipage}[t]{\linewidth}
\vspace{2mm}
\begin{verbatim}
def func_0(self, value, list_num, index):
    self.value = value
    self.list_num = list_num
    self.index = index
def func_1(self, other):
    return self.value < other.value
\end{verbatim}
\vspace{2mm}
\centering \textcolor{RerankGreen}{\textbf{Code Score:} 0.5446}
\vspace{2mm}
\end{minipage} & 
\begin{minipage}[t]{\linewidth}
\vspace{2mm}
This code implements a binary search algorithm to find the maximum subarray sum within a given list of arrays. It uses a priority queue to efficiently find the maximum and minimum values in the subarrays. The algorithm iteratively selects the subarray with the maximum sum, updates the maximum and minimum values, and repeats until the end of the subarray is reached.

\vspace{3mm}
\centering \textcolor{BaselineRed}{\textbf{Description Score:} 0.5057}
\vspace{2mm}
\end{minipage} \\
\hline
\multicolumn{2}{|p{\textwidth}|}{
\vspace{1mm}
\centering\textbf{Combined Retriever Scores ($\boldsymbol{\alpha}$=0.7):} \textcolor{RerankGreen}{Positive Example: \textbf{0.5882}}, \textcolor{BaselineRed}{Negative Example: 0.5174}
\vspace{1mm}
} \\
\hline
\end{tabular}
\caption{Case Study 2 (HumanEval/68): The baseline incorrectly ranks code for binary search higher than the correct solution for finding the smallest even number. Despite lower code similarity, semantic-augmented reranking correctly prioritizes the ground truth by leveraging description similarity.}
\label{tab:case_study_2}
\end{table*}

\begin{table*}
\small
\centering
\caption{Case studies showing how semantic-augmented localization helps identify the correct files for modification which leads to correct patches being generated.}
\begin{tabular}{p{4cm}p{4.5cm}p{6.5cm}}
\toprule
\multicolumn{3}{c}{\textbf{django-11964:} "The value of a TextChoices / IntegerChoices field has a differing type"} \\
\midrule
\textbf{Repository Structure} & \textbf{Localized Files} & \textbf{Patch Snippet} \\
\midrule
\textbf{Baseline:} \newline
django/ \newline
$\mid$ \quad $\ldots$ \newline
$\mid$ \quad db/ \newline
$\mid$ \quad $\mid$ \quad $\ldots$ \newline
$\mid$ \quad $\mid$ \quad models/ \newline
$\mid$ \quad $\mid$ \quad $\mid$ \quad $\ldots$ \newline
$\mid$ \quad $\mid$ \quad $\mid$ \quad enums.py
\newline\newline
\textbf{\modelname method:} \newline
django/ \newline
$\mid$ \quad $\ldots$ \newline
$\mid$ \quad db/ \newline
$\mid$ \quad $\mid$ \quad $\ldots$ \newline
$\mid$ \quad $\mid$ \quad models/ \newline
$\mid$ \quad $\mid$ \quad $\mid$ \quad $\ldots$ \newline
$\mid$ \quad $\mid$ \quad $\mid$ \quad \texttt{enums.py} \newline
\quad \textit{{\small - This Python file provides a custom implementation of enums with additional features such as ...}}
&
\textbf{Baseline:} \newline
\textcolor{red}{django/contrib/admin/options.py} \newline
\textcolor{red}{django/.../templates.py} \newline
\textcolor{red}{django/db/models/fields/\_\_init\_\_.py} 
\newline\newline
\textbf{\modelname method:} \newline
django/db/models/fields/\_\_init\_\_.py \newline
\textcolor{green}{django/db/models/enums.py} \newline
django/db/models/base.py 
&
\textbf{Baseline (wrong file):} 
\begin{verbatim}
# in django/db/models/fields/__init__.py
- return getattr(obj, self.attname)
+ value = getattr(obj, self.attname)
+ if isinstance(value, enum.Enum):
+   return value.value
+ return value
\end{verbatim}
\textbf{\modelname method (correct file):}
\begin{verbatim}
# in django/db/models/enums.py
class Choices(enum.Enum, metaclass=...):
-  pass
+  def __str__(self):
+    return str(self.value)
\end{verbatim} 
\\
\midrule
\multicolumn{3}{c}{\textbf{sympy-19487:} "Rewrite the sign function in terms of Abs in SymPy"} \\
\midrule
\textbf{Repository Structure} & \textbf{Localized Files} & \textbf{Patch Snippet} \\
\midrule
\textbf{Baseline:} \newline
sympy/ \newline
$\mid$ \quad $\ldots$ \newline
$\mid$ \quad functions/ \newline
$\mid$ \quad $\mid$ \quad $\ldots$ \newline
$\mid$ \quad $\mid$ \quad elementary/ \newline
$\mid$ \quad $\mid$ \quad $\mid$ \quad $\ldots$ \newline
$\mid$ \quad $\mid$ \quad $\mid$ \quad complexes.py
\newline\newline
\textbf{\modelname method:} \newline
sympy/ \newline
$\mid$ \quad $\ldots$ \newline
$\mid$ \quad functions/ \newline
$\mid$ \quad $\mid$ \quad $\ldots$ \newline
$\mid$ \quad $\mid$ \quad elementary/ \newline
$\mid$ \quad $\mid$ \quad $\mid$ \quad $\ldots$ \newline
$\mid$ \quad $\mid$ \quad $\mid$ \quad \texttt{complexes.py} \newline
\quad \textit{{\small - This Python file defines several mathematical functions for symbolic computation using the SymPy library ...}}
&
\textbf{Baseline:} \newline
\textcolor{red}{sympy/core/expr.py} \newline
\textcolor{red}{sympy/core/function.py} \newline
\textcolor{red}{sympy/.../miscellaneous.py} \newline
\textcolor{red}{...} 
\newline\newline
\textbf{\modelname method:} \newline
sympy/core/function.py \newline
\textcolor{green}{sympy/.../complexes.py} \newline
sympy/.../miscellaneous.py \newline
... 
&
\textbf{Baseline (wrong file):}
\begin{verbatim}
# in sympy/core/function.py 
+ def _eval_rewrite_as_Abs(self, *args, **):
+   from sympy import Abs
+   if len(args) == 1:
+     arg = args[0]
+     if arg.is_zero:
+       return 0
+     return arg / Abs(arg)
\end{verbatim}
\textbf{\modelname method (correct file):}
\begin{verbatim}
# in sympy/.../complexes.py 
class sign(Function):
+  def evalrewrite_as_Abs(self, arg, **):
+    if arg.is_zero:
+      return S.NaN
+    return arg / Abs(arg)
\end{verbatim} 
\\
\bottomrule
\end{tabular}
\label{tab:case_studies_loc}
\end{table*}

\end{document}